\documentclass{Interspeech}

\interspeechcameraready
\usepackage{caption}
\usepackage{supertabular} 
\usepackage[utf8]{inputenc}
\usepackage{CJKutf8, multirow}
\usepackage[colorlinks,linkcolor=blue,citecolor=blue,urlcolor=blue]{hyperref}

\title{Dolphin: A Large-Scale Automatic Speech Recognition Model for Eastern Languages}

\author[affiliation={1}]{Yangyang}{Meng$^*$}
\author[affiliation={2}]{Jinpeng}{Li$^*$}
\author[affiliation={2}]{Guodong}{Lin$^*$}
\author[affiliation={2}]{Yu}{Pu$^*$}
\author[affiliation={1}]{Guanbo}{Wang$^*$}
\author[affiliation={1}]{Hu}{Du$^*$}
\author[affiliation={1}]{Zhiming}{Shao$^{\dagger}$}
\author[affiliation={1}]{Yukai}{Huang}
\author[affiliation={1}]{Ke}{Li}
\author[affiliation={2}]{Wei-Qiang}{Zhang$^{\dagger}$}

\address{
  $^1$Dataocean AI \\
  $^2$Speech and Audio Technology Lab, Dept. EE, Tsinghua University
  }
\email{wqzhang@tsinghua.edu.cn}


\usepackage{comment}

\begin{document}
\begin{CJK}{UTF8}{gbsn}
\maketitle

\begingroup
\renewcommand{\thefootnote}{}
\footnotetext{$^*$ Co-first authors, equal contribution}
\footnotetext{$^{\dagger}$ Corresponding author}
\endgroup

\begin{abstract}
This report introduces Dolphin, a large-scale multilingual automatic speech recognition (ASR) model that extends the Whisper architecture to support a wider range of languages. Our approach integrates in-house proprietary and open-source datasets to refine and optimize Dolphin’s performance. The model is specifically designed to achieve notable recognition accuracy for 40 Eastern languages across East Asia, South Asia, Southeast Asia, and the Middle East, while also supporting 22 Chinese dialects. Experimental evaluations show that Dolphin significantly outperforms current state-of-the-art open-source models across various languages. To promote reproducibility and community-driven innovation, we are making our trained models and inference source code publicly available.
\end{abstract}

\section{Introduction}

Automatic Speech Recognition (ASR) has witnessed remarkable progress in recent years, driven by advances in model architectures and the availability of large-scale datasets. Some notable datasets have contributed to this progress \cite{panayotov2015librispeech, hernandez2018ted, ardila2019common, pratap2020mls, wang2021voxpopuli, chen2021gigaspeech, zhang2022wenetspeech, conneau2023fleurs}, which provides a large, multilingual collection of speech data. In terms of ASR architectures, significant strides have been made with models such as Self-Supervised Learning (SSL)-based architectures \cite{baevski2020wav2vec,sadhu2021wav2vec-c,hsu2021hubert,chung2021w2v,chen2022wavlm,chiu2022self,baevski2022data2vec,baevski2022efficient,zhang2023google,ZhaoJ2022JSTSP}, which rely on large amounts of unlabeled data, Large Language Model (LLM)-based architectures \cite{zhang2023speechgpt, chu2023qwen, xie2024mini}, which incorporate deep contextual understanding, and supervised learning models including traditional structures based on Deep Neural Networks (DNNs) \cite{Hinton2012DeepNN, 10.1109/TASL.2011.2134090}, Convolutional Neural Networks (CNNs) \cite{6857341}, and Recurrent Neural Networks (RNNs) \cite{sak2014longshorttermmemorybased} as well as end-to-end architectures, such as Listen, Attend and Spell (LAS) \cite{chan2015listenattendspell}, Deep Speech2 \cite{Amodei2015DeepS2}, Conformer \cite{Gulati2020ConformerCT}, Whisper \cite{radford2023robust}, Paraformer \cite{gao2023paraformerfastaccurateparallel} and so on.

Among prominent speech recognition models, Whisper has gained widespread recognition due to its outstanding multilingual capabilities, robust performance across diverse languages, and accessibility to the research community. Through large-scale training on a 680,000-hour multilingual corpus, Whisper has established new benchmarks in ASR accuracy, particularly for Western languages. However, two critical limitations hinder its broader application.

First, reproducibility challenges persist despite the model's open-source nature. While the architecture and inference code are publicly accessible, the proprietary training pipeline and undisclosed data curation methods prevent full replication of reported results. Second, performance disparities emerge in cross-linguistic comparisons: our preliminary analysis shows a significant performance difference between Western and Eastern language languages.

In response to these challenges, the research community has pursued two complementary directions. The Open Whisper-style Speech Model (OWSM) initiative \cite{peng2023reproducing, peng2024owsm} has developed fully reproducible architectures with transparent training protocols. Concurrently, significant dataset efforts \cite{ardila2019common, zhang2022wenetspeech, li2023yodas, yang2024gigaspeech, yin2023reazonspeech, bang2020ksponspeech, openstt} have expanded linguistic coverage for Eastern languages through curated corpora.

\begin{table*}[h]
  \caption{Details of data, model architectures, and training configurations.}
  \label{tab:model_param}
  \centering
  \scalebox{0.95}{
  \begin{tabular}{lccccccccc}
    \toprule
    \multirow{2}{*}{} & \multicolumn{5}{c}{\textbf{Whisper}}  & \multicolumn{3}{c}{\textbf{Dolphin}} \\
    ~& \textbf{base} & \textbf{small} & \textbf{medium} & \textbf{large-v1} & \textbf{large-v3} & \textbf{base} & \textbf{small} & \textbf{medium} & \textbf{large}\\
    \midrule
    \multicolumn{9}{l}{\textbf{Network architecture}} \\
    \quad Parameters&74M&244M&769M&1550M&1550M&140M&372M&910M&1679M\\
    \quad Encoder& \multicolumn{4}{c}{Transformer} & Transformer & \multicolumn{4}{c}{E-Branchformer}\\
    \quad Decoder& \multicolumn{4}{c}{Transformer} & Transformer & \multicolumn{4}{c}{Transformer}\\
    \quad Layers & 6&12&24&32&32&6&12&16&20\\
    \quad Hidden size & 512	& 768&1024&1280&1280 & 512&768&1024&1280\\
    \quad Attention heads&8&12&16&20&20&8&12&16&20\\
    \quad Time shift&20ms&20ms&20ms&20ms&20ms&40ms&40ms&40ms&40ms\\
    \midrule
    \multicolumn{9}{l}{\textbf{Training data}} \\
    \quad Unlabelled hours & \multicolumn{4}{c}{-} & 4M & \multicolumn{4}{c}{-} \\
    \quad Labelled hours & \multicolumn{4}{c}{680k} & 1M & \multicolumn{4}{c}{212k} \\
    \quad Languages & \multicolumn{4}{c}{99} & 100 & \multicolumn{4}{c}{40} \\
    \quad BPE vocabulary & \multicolumn{4}{c}{52k} & 52k & \multicolumn{4}{c}{40k} \\
    \midrule 
    \multicolumn{9}{l}{\textbf{Hyperparameters}} \\
    \quad Batch size & \multicolumn{4}{c}{256} & unknown & \multicolumn{4}{c}{1024} \\
    \quad Total updates & \multicolumn{4}{c}{1M} & 2epochs & \multicolumn{4}{c}{4epochs} \\
    \quad Warmup updates & \multicolumn{4}{c}{2k} & unknown & \multicolumn{4}{c}{2k} \\
    \quad Learning rate & 1e-3&5e-4&2.5e-4&1.75e-4&unknown & 5e-4&5e-4&2.5e-4&2e-4 \\
    \quad Optimizer & \multicolumn{4}{c}{AdamW} & unknown & \multicolumn{4}{c}{AdamW} \\
    \quad CTC weight & \multicolumn{4}{c}{-} & - & \multicolumn{4}{c}{0.3} \\
    \bottomrule
  \end{tabular}
  }
\end{table*}

Building upon these foundations, we present Dolphin---a large-scale multilingual and multitask ASR model. Dolphin focuses on optimizing performance for Eastern languages, offering significant improvements over existing state-of-the-art (SOTA) systems.


Our work presents several highlights and contributions:

\begin{itemize}

\item Dolphin closes the performance gap between Eastern and Western languages in ASR, achieving recognition accuracy for Eastern languages that is on par with its performance for Western languages. This accomplishment is enabled through a training pipeline that integrates proprietary internal data and publicly available datasets. Comprehensive evaluations using both in-domain and out-of-domain test sets ensure the model's robust generalization capabilities.


\item When comparing models of the same size (base, small, medium, and large), Dolphin consistently outperforms Whisper across three diverse test sets. Notably, Dolphin base, small, and medium models achieve comparable performance to Whisper large-v3 model, demonstrating the effectiveness of Dolphin’s architecture and training approach.


\item Across three test sets, the Dolphin small model shows an average 24.5\% improvement in Word Error Rate (WER) compared to the base model, the medium model achieves an additional 8.3\% improvement over the small model, and the large model achieves an additional 6.5\% improvement over the medium model. These results align with the Scaling Law, suggesting that larger Dolphin models can potentially achieve SOTA performance across a wider range of languages.


\item We are releasing the Dolphin base and small models along with the inference code\footnote{\url{https://github.com/DataoceanAI/Dolphin}}. This initiative is expected to set a strong foundation for future research and open-source community. 

\end{itemize}

With these contributions, Dolphin represents a significant step forward in addressing the challenges of multilingual ASR, particularly for Eastern languages, facilitating further advancements in multilingual ASR technology, and promoting innovation in the field.

\section{Methods}


\subsection{Model Architecture}

Following the approach outlined in OWSM \cite{peng2024owsm}, we adopt a joint CTC-Attention architecture \cite{kim2017joint, watanabe2017hybrid}, which combines the advantages of both Connectionist Temporal Classification (CTC) and Attention-based mechanisms. This hybrid approach enables robust and efficient training for large-scale multilingual speech recognition.

The encoder in our architecture is based on E-Branchformer \cite{kim2023branchformer}, a state-of-the-art model that incorporates parallel branch structures. This design allows the model to capture both local and global dependencies in the input speech signals more effectively. For the decoder, we employ the standard Transformer \cite{vaswani2017attention}, which has proven to be effective in sequence-to-sequence tasks.

To further improve training efficiency and performance, we incorporate $4\times$ subsampling layer, which reduces the sequence length of the input features and accelerates computation. Detailed model parameters are shown in Table \ref{tab:model_param}.

We train four sizes of models, corresponding to Whisper base, small, medium, and large models, respectively. In each scale, Dolphin has slightly more parameters than Whisper, due to E-Branchformer encoder, CTC layer and subsampling layer. 


\begin{figure*}[t]
  \centering
  \includegraphics[width=\textwidth]{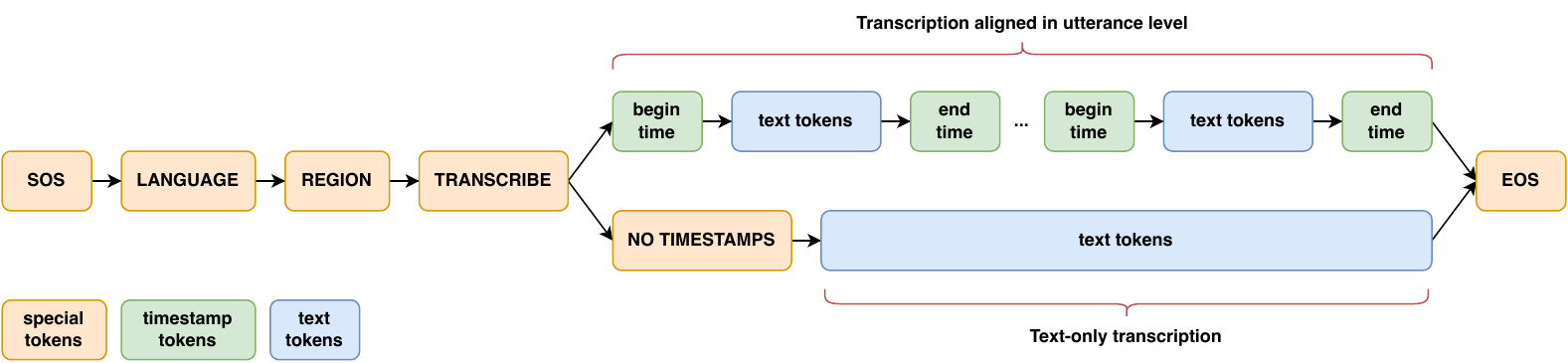}
  \caption{Multitask format used by Dolphin, which mostly follows OpenAI Whisper\cite{radford2023robust}. Dolphin focuses on ASR and does not support translation task. In addition, Dolphin introduces region-specific tokens, thus enabling support for dialects.}
  \label{fig:multitask-data-format}
\end{figure*}

\subsection{Multitask Format}

Whisper creatively introduced a sequence-to-sequence architecture that leverages a flexible token-based design to support a wide variety of speech-related tasks, including transcription, translation, voice activity detection (VAD), segmentation, and language identification (LID). This design enables a single model to handle multiple tasks efficiently by utilizing task-specific tokens to guide the model's behavior.

Dolphin largely follows this innovative design approach of Whisper and OWSM, but introduces several key modifications for its specific focus on ASR. Dolphin does not support translation tasks, and eliminates the use of previous text and its related tokens. These simplify the input format and reduce potential complexity.

A significant enhancement in Dolphin is the introduction of a two-level language token system to better handle linguistic and regional diversity, especially in Dataocean AI dataset. The first token specifies the language (e.g., \texttt{<zh>}, \texttt{<ja>}), while the second token indicates the region (e.g., \texttt{<CN>}, \texttt{<JP>}). This hierarchical approach allows the model to capture differences between dialects and accents within the same language, as well as similarities across languages within the same region. Our design improves the model's ability to distinguish between closely related dialects and enhances its generalization capabilities by establishing connections between languages and regions.  This is particularly beneficial in a multilingual and multi-dialectal context. Additionally, Dolphin's multitask format enables the model to explicitly recognize accent and dialect, in both speech recognition and language identification. Figure~\ref{fig:multitask-data-format} illustrates the multitask data format used in Dolphin, highlighting the integration of language token and region token. 



\section{Training Data}

In constructing our dataset, we focused on Eastern languages, recognizing the potential for shared linguistic features among them. This choice is strategically significant, as training multilingual models with a foundation in Eastern languages can enhance performance across various dialects, fostering better communication among people in Eastern nations. Furthermore, a secondary motivation for selecting Eastern languages stems from the observed limitations of existing multilingual speech recognition models, such as Whisper, which often underperform in accurately processing these languages. To address this gap, we leveraged a combination of internal data alongside publicly available datasets, resulting in the creation of a robust collection comprising over 200,000 hours of audio. This extensive dataset serves as a solid foundation for achieving high-performance outcomes in our models. Additionally, we standardized the dataset by implementing a consistent data format and introducing specialized labels, including language, region, and task identifiers. This facilitates both timestamped and non-timestamped entries, as well as options for punctuation, ensuring comprehensive coverage of the diverse characteristics inherent in data.

\subsection{Datasets}

\begin{table}[t]
    \centering
    \caption{Dataset Statistics After Cleaning}
    \begin{tabular}{l r l l}
        \hline
        \textbf{Dataset}      & \textbf{Duration (h)} & \textbf{Language}   & \textbf{Source}      \\
        \hline
        Dataocean AI     & 137,712  & Multilingual  & Proprietary  \\
        ReazonSpeech & 35,000   & Japanese      & Open Source  \\
        GigaSpeech2  & 22,015   & Multilingual  & Open Source  \\
        WenetSpeech  & 10,000   & Chinese       & Open Source  \\
        Yodas        & 5,981    & Multilingual  & Open Source  \\
        OpenSTT      & 5,727    & Russian       & Open Source  \\
        KsponSpeech  & 969      & Korean        & Open Source  \\
        CommonVoice  & 733      & Multilingual  & Open Source  \\
        \hline
        \textbf{Total} & \textbf{212,137} & - & - \\
        \hline
    \end{tabular}
    \label{tab:dataset_stats}
\end{table}

\subsubsection{Dataocean AI Dataset}

Dataocean AI dataset is an internally curated, high-quality dataset developed by Dataocean AI. This dataset is an integration of our vast, high-quality commercial dataset collections, encompassing a total of 137,712 hours of audio across 38 Eastern languages. Additionally, it includes 22 Chinese dialects (see Appendix \ref{sec:appendixb} for the full list). The dataset is carefully annotated and covers a wide variety of languages, scenarios, and contexts, ensuring diversity and richness in the data. This broad coverage allows for comprehensive model training, with a focus on Eastern languages. We primarily utilize it in our experiments as the main source of training data.

\subsubsection{Open Source Datasets}

In addition to our internal dataset, we incorporate the following widely accessible open-source datasets to enhance the diversity and robustness of our research:

\begin{itemize}

    \item \textbf{Common Voice} \cite{ardila2019common} is a multilingual open-source speech dataset. It includes contributions from volunteers in a variety of languages, covering different accents, dialects, and speaking styles. We include 733 hours in 29 languages from it.

    \item \textbf{YODAS} \cite{li2023yodas} (YouTube-Oriented Dataset for Audio and Speech) is a large-scale, multilingual dataset sourced from YouTube videos. We incorporate 5,981 hours across 3 languages from it.

    \item \textbf{GigaSpeech 2} \cite{yang2024gigaspeech} is a large-scale speech recognition dataset focusing on Southeast Asian languages, covering Thai, Indonesian, and Vietnamese. We use the GigaSpeech 2 refined which consists of 22,015 hours of speech data.

    \item \textbf{WenetSpeech} \cite{zhang2022wenetspeech} is a Mandarin Chinese speech dataset containing 10,000 hours of speech data.

    \item \textbf{ReazonSpeech} \cite{yin2023reazonspeech} is a Japanese speech recognition dataset, which includes 35,000 hours of speech data extracted from news programs and broadcasts, covering various dialects and accents.

    \item \textbf{KsponSpeech} \cite{bang2020ksponspeech} is a Korean speech recognition dataset, which contains over 1,000 hours of speech data, covering various scenarios and topics such as news, interviews, and daily conversations. 

    \item \textbf{OpenSTT} \cite{openstt} is a Russian speech recognition dataset which includes about 20,000 hours of speech data extracted from news programs, broadcasts, and movies, covering various dialects and accents. We include 5,727 hours from it.

\end{itemize}

\begin{figure*}[t]
  \centering
  \includegraphics[width=\linewidth]{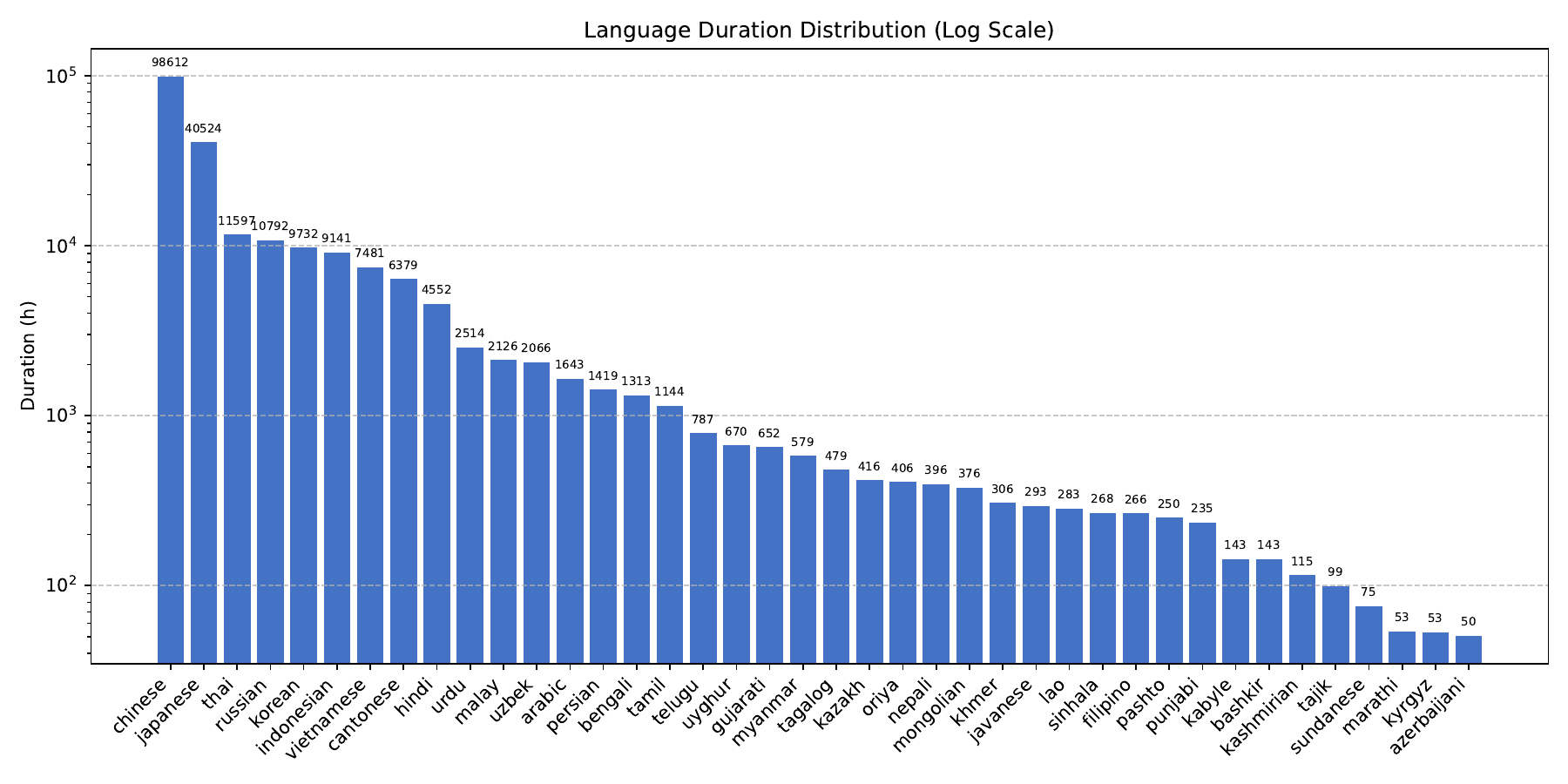}
  \caption{The distribution of data duration across 40 Eastern languages in the cleaned dataset, represented on a logarithmic scale. There are 36 languages with a data duration greater than 100 hours, and 16 languages with a data duration exceeding 1000 hours.}
  \label{fig:duration-language}
\end{figure*}

\subsection{Data Processing}
\subsubsection{Data Cleaning}
Unlike Whisper, whose training data primarily consists of audio-text pairs sourced from the internet, our training dataset comprises proprietary data from Dataocean AI and publicly available open-source datasets. For datasets such as YODAS, which contain human-annotated and ASR-generated transcriptions, we exclusively use the human-annotated portion. As a result, most of our training data is manually transcribed, ensuring a higher transcription quality. We believe that this data quality, particularly the quality of transcriptions, is a key factor enabling our model to achieve significantly better recognition performance than Whisper, even with a smaller model size.

The cleaned text format largely aligns with Whisper’s conventions. Before the transcription content, metadata tags indicate information such as language, task type, punctuation presence, timestamp inclusion, and whether the text contains non-standardized elements (e.g., Arabic numerals). Given that certain languages exhibit notable pronunciation or annotation differences across regions, we adopt a more granular language tagging approach based on the BCP 47 language tag standard \cite{rfc5646}. We refer to this as secondary language tagging, which includes both the language and regional identifiers. For example, \verb|<ru><RU>| represents Russian in Russia, while \verb|<ru><BY>| denotes Russian in Belarus.

For timestamps, we employ the same sentence-level timestamping approach as Whisper, where timestamp tokens mark the start and end of each sentence. For long audio recordings (typically several minutes in length), we segment them into smaller clips during data preprocessing and later merge them into long-form audio sequences.

After formatting the text, we perform statistical analysis and filtering to ensure data quality. This includes measuring text similarity before and after cleaning, validating timestamps and punctuation accuracy, and computing the text-to-speech ratio (i.e., the ratio of text length to speech duration) to discard data with excessively high ratios. We convert all training audio into the standardized .wav format to enhance audio processing efficiency.

\subsubsection{Training Data Processing}
During training, we explored various data processing strategies.

In the initial version of the training data, we directly used the cleaned dataset. However, a major issue was the high proportion of short-duration audio samples. Most audio clips were around 5 seconds long, leading to an excessive deletion error rate across multiple languages. This issue was consistent with the fact that most of the training data consisted of short audio samples.

To address this, we experimented with an alternative approach by concatenating the cleaned audio data into longer segments of 25–30 seconds. This significantly mitigated the high deletion error rate. While this approach resulted in a slight increase in insertion errors, the overall recognition performance improved, leading to an average WER reduction of 9.01\%.


Building on this improvement, we further reorganized the raw data into six different length-based buckets to better balance duration distribution. The resulting data duration distribution is shown below. With this refined approach, insertion errors were reduced for most languages, contributing to an additional 5.03\% reduction in average WER.

\section{Experiments and Results}

\begin{table}[t]
  \caption{Performance of Dolphin models on various datasets. The evaluation metric is WER (\%).}
  \label{tab:model_dosm}
  \centering
  \scalebox{0.87}{
  \begin{tabular}{lc|cccc}
    \toprule
    \multirow{2}{*}{\textbf{Dataset}} & \textbf{Supported} & \multicolumn{4}{c}{\textbf{Dolphin}} \\
    ~&\textbf{Languages}& \textbf{base} & \textbf{small} & \textbf{medium} & \textbf{large} \\
    \midrule
    Dataocean AI & 38 & 31.5 & 24.5 &22.2 & 21.4 \\
    Fleurs & 33 & 31.2 & 23.6 & 22.2 & 20.6  \\
    CommonVoice & 29 & 37.2 & 27.4 & 25.0 & 22.9 \\
    \midrule
   average & - & 33.3 & 25.2 &23.1 & 21.6 \\
    \bottomrule
  \end{tabular}
  }
\end{table}

\subsection{Experimental Setup}

\begin{table*}[t]
  \caption{Performance comparison of OpenAI Whisper and Dolphin models on various datasets. The evaluation metric is WER (\%).}
  \label{tab:model_comparison}
  \centering
  \scalebox{0.95}{
  \begin{tabular}{lc|cccc|cccc}
    \toprule
    \multirow{2}{*}{\textbf{Dataset}} & \multirow{2}{*}{\textbf{Intersect Languages}} & \multicolumn{4}{c}{\textbf{Whisper}}  & \multicolumn{4}{c}{\textbf{Dolphin}} \\
    ~&~& \textbf{base} & \textbf{small} & \textbf{medium} & \textbf{large-v3} & \textbf{base} & \textbf{small} & \textbf{medium} & \textbf{large} \\
    \midrule
    Dataocean AI & 32 & 88.6 & 79.3 & 71.8 & 57.8 & 29.4 & 22.4 &20.3 & 19.4 \\
    Fleurs & 29 & 82.3 & 69.8 & 62.1 & 48.8 & 30.4 & 23.0 & 21.1  &19.8 \\
    CommonVoice & 24 & 87.4 & 77.1 & 70.4 & 50.2 & 35.6 & 26.6 & 24.5 & 22.5 \\
    \midrule
   average & - & 86.1 & 75.4 & 68.1 & 52.3 & 31.8 & 24.0 &22.0 & 20.6 \\
    \bottomrule
  \end{tabular}
  }
\end{table*}

We extract 80-channel log Mel-scale Filter Bank energies (fbank) as input feature, with a frame length of 25 ms and a frame shift of 10 ms. We apply SpecAugment \cite{park2019specaugment} and global normalization to make models more robust. BPE \cite{sennrich2015neural} vocabulary size is 40K. The model hyperparameters are presented in Table \ref{tab:model_param}. 

During the training, we utilize AdamW optimizer \cite{loshchilov2017decoupled}. Regularization methods include Dropout \cite{srivastava2014dropout} rate of 0.1, layer normalization \cite{ba2016layer}, and label smoothing \cite{szegedy2016rethinking}. The weight of CTC loss is set to 0.3, and 50\% of training batches are padded to 30 s.  The learning rate schedule consists of a linear warmup for the first 2K steps, followed by exponential decay after reaching the peak. Models are trained for 4 epochs on 16 NVIDIA H100 GPUs, with a batch size of 1024. Dolphin base, small, medium and large are trained for approximately 2, 5, 10 and 18 days, respectively. 

During inference, the final evaluated checkpoint is obtained by averaging 30 to 40 checkpoints. Outputs are only from attention decoder and CTC layers are not used. Each utterance is padded to 30 s before decoding. We set beam size to 5, and maxlen ratio to 0.5. Time stamps are predicted but discarded during evaluation. 

\subsection{Technical Challenges}

\subsubsection{Memory Footprint Issue}

\begin{figure}[t]
  \centering
  \includegraphics[width=\linewidth]{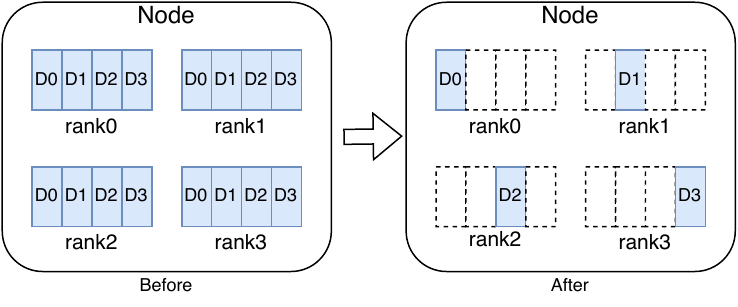}
  \caption{Data loading strategy optimization. Assume a node with 4 GPUs, each GPU is assigned a corresponding process, referred to as a rank. Before optimization, each rank loads a complete copy of the dataset, denoted as \{D0,D1,D2,D3\}. After optimization, each rank is assigned only the subset of the dataset it requires for computation.}
  \label{fig:memory-footprint-optimization}
\end{figure}

Our training set consists of 160 million utterances, an Out of Memory (OOM) issue was encountered during the data processing phase.

We conducted an in-depth analysis of the sampler, dataset, and dataloader modules for data processing and found that the large number of utterances caused memory overflow. PyTorch supports two types of datasets: map-style and iterable-style. ESPnet \cite{watanabe2018espnet} uses the map-style. The map-style dataset loads the utterance metadata (the mapping between utterance id and text, audio) into memory. The memory footprint grows linearly with the number of training data utterances. To improve data loading speed, there will be multiple workers inside the dataloader for data prefetching, which further increases the memory footprint of the physical machine. Eventually lead to OOM.

Inspired by the Zero-DP \cite{rajbhandari2020zero}, we propose a data sharding strategy in Figure \ref{fig:memory-footprint-optimization} . Instead of loading the entire dataset replication, each rank is optimized to only load the necessary subset of the dataset. This approach significantly reduces the memory footprint of each rank, thereby leading to a reduction in the overall memory consumption on the physical machine. Furthermore, as the degree of data parallelism increases, the memory footprint of a single node decreases linearly.

\subsubsection{Training Efficiency}

Merging short audios into long audio can significantly increase the computational density and utilization of the GPU, thus significantly improving the training efficiency.

In our dataset, audio duration exhibits a significant left-skewed distribution, with a high proportion of short audio (1-10 seconds) and a low proportion of long audio (11-30 seconds). To achieve a more balanced distribution of audio durations, we merged short audios and redistributed them evenly into 5-second interval buckets within the 0-30 second range.

When processing an large-scale dataset of 210,000 hours, using ffmpeg to physically merge multiple short audios into longer audio would be highly time-intensive. Instead, we adopted a more efficient logical merging strategy. Specifically, during the data preparation phase, we use a dictionary to represent the mapping relationship before and after audio merging and dynamically merged audios during training.

 With the optimized merging strategy, the training time for a single epoch of the small model was significantly reduced from 64 hours to 28.6 hours, achieving a 123.78\% increase in training speed. This improvement greatly accelerated the model iteration process.

\subsection{Results}

Table \ref{tab:model_dosm} presents the number of supported languages and the average WER across three multilingual test sets for four model sizes of Dolphin: base, small, medium and large. The average WERs for the base, small, medium, and large models are 33.3\%, 25.2\%, 23.1\% and 21.6\%, respectively, demonstrating a practical capability for real-world applications. Notably, the small model is approximately 2.7 times larger than the base model, featuring 341 million parameters and 31 million CTC parameters. In comparison to the base model, the small model achieves a relative reduction in WER of about 24.3\%, indicating significant improvement and making it an excellent choice for balancing scale and performance. Furthermore, the medium model is approximately 2.4 times the size of the small model, yielding a relative WER reduction of approximately 8.3\% compared to the small model, the large model is approximately 1.8 times the size of the medium model, yielding a relative WER reduction of approximately 6.5\% compared to the medium model.

Table \ref{tab:model_comparison} shows a comparison of the performance of the OpenAI Whisper model and our proposed Dolphin model in various intersection-supported languages on a multilingual test set evaluated on the metric of WER. The results show that the Dolphin model consistently outperforms Whisper on the Oriental languages dataset as a whole, with significant decreases in WER compared to Whisper on multiple languages, and the Appendix \ref{sec:appendixa} shows detailed results for each language. When the model parameters are comparable, our Dolphin model demonstrates a significant improvement over Whisper in terms of average performance on intersect languages. For the base, small, medium and large models of both systems, the relative reductions in WER for Dolphin compared to Whisper are approximately 63\%, 68\%, 68\% and 61\%, respectively.
It is noteworthy that even the base model of Dolphin achieves an impressive WER, significantly lower than the Whisper large-v3 model, on all datasets. For example, referring to the average of the three multilingual datasets, the WER of Dolphin's base model was 31.8\%, while Whisper large-v3 has a WER of 52.3\%, highlighting the performance advantage. From this perspective, the WER of the Dolphin base model is comparatively reduced by 39\% when assessed against the Whisper large-v3 model for these languages, despite the fact that the size of the Dolphin model is less than 1/10 that of Whisper large-v3. The WERs of Dolphin's small, medium and large models are further reduced to 24.0\%, 22.0\% and 20.6\%, respectively. On the DataoceanAI, Fleurs, and CommonVoice test sets, compared to the Whisper large-v3 model, the Dolphin medium model achieved an average WER reduction of 65\%, 57\% and 51\%, respectively, while the Dolphin large model achieved an average WER reduction of 66\%, 59\%, and 55\%.

Table \ref{tab:model_dosm_dialect} presents the performance of the Dolphin models on Chinese dialects. We evaluated the model on the KeSpeech test set \cite{tang2021kespeech}, which consists of one Mandarin subset and eight Chinese dialect subsets. These results demonstrate progressive performance improvements as the model size increases, highlighting the capability of handling Chinese dialects effectively.

The results of the study demonstrate two key points. Firstly, our hybrid training approach combining proprietary and open source data achieves superior cross-language generalisation capabilities, demonstrating the reliability and excellence of our dataset. Second, we optimised the architecture inherited from OWSM to provide strong multilingual modelling capabilities and better performance for training.

\begin{table}[t]
  \caption{Performance of Dolphin models on Chinese accent testset. The evaluation metric is CER (\%).}
  \label{tab:model_dosm_dialect}
  \centering
  \begin{tabular}{l|cccc}
    \toprule
    \multirow{2}{*}{\textbf{Dataset}} & \multicolumn{4}{c}{\textbf{Dolphin}} \\
    ~& \textbf{base} & \textbf{small} & \textbf{medium} & \textbf{large} \\
    \midrule
    KeSpeech & 14.75 & 10.94 & 10.17 & 9.23 \\
    \bottomrule
  \end{tabular}
\end{table}

\section{Future Work}

While Dolphin has demonstrated significant advancements in ASR for Eastern languages and across multiple test sets, several areas remain for future exploration and improvement:

\begin{itemize}
\item Inspired by the observed performance gains with larger model sizes, we will focus on training and evaluating larger Dolphin models. These models are expected to achieve state-of-the-art results across an even broader set of languages, further enhancing Dolphin’s generalization and performance.

\item While we currently focus on Eastern languages, we will also aim to broaden language coverage, particularly for underrepresented and low-resource languages. This will include curating additional datasets and optimizing training strategies for these languages.

\item To address real-world deployment scenarios, we plan to optimize Dolphin for low latency and real-time performance while maintaining accuracy. This includes refining the model architecture, implementing efficient inference techniques, and compression.

\end{itemize}

\section{Conclusion}

In this report, we have presented Dolphin, a large-scale multilingual multitask ASR model. Built upon the Whisper-style architecture and based on OWSM, Dolphin integrates proprietary and publicly available datasets. Experimental results show that Dolphin consistently outperforms existing SOTA models across a wide range of languages and model sizes, effectively bridging the performance gap between Eastern and Western languages. In particular, Dolphin base model outperforms Whisper large-v3. 
Through the open-source release of Dolphin base and small models, along with inference code, we aim to contribute to further advancements in multilingual speech processing. 


\bibliographystyle{IEEEtran}
\bibliography{tech_report}
\end{CJK}

\clearpage

\appendix

\onecolumn
\section{Detailed Results for each Language}

\label{sec:appendixa}
\begin{table*}[h]
  \caption{WER (\%) on Dataocean AI dataset.}
  \label{tab:dataocean results}
  \centering
  \scalebox{1}{
  \begin{tabular}{l|cccc|cccc}
    \toprule
     & \multicolumn{4}{c}{\textbf{Whisper}} & \multicolumn{4}{c}{\textbf{Dolphin}} \\
    \textbf{Models} & \textbf{base} & \textbf{small} & \textbf{medium}  & \textbf{large-v3} & \textbf{base} & \textbf{small} & \textbf{medium} & \textbf{large} \\
    \midrule
    Chinese & 53.1 & 43.2 & 34.3 & 27.9 & 11.8 & 9.7 & 9.2  & 9.0 \\
    Japanese & 41.5 & 26.9 & 25.1 & 19.5 & 18.8 & 18.3 & 13.8 & 14.0 \\
    Thai & 38.3 & 25.5 & 18.3 & 10.9 & 7.0 & 5.7 & 5.4  & 5.2\\
    Russian & 49.7 & 40.5 & 38.1 & 29.4 & 38.2 & 34.2 & 31.2 & 32.3 \\
    Korean & 32.5 & 21.7 & 21.1 & 14.1 & 20.1 & 11.6 & 13.0 & 11.5\\
    Indonesian & 46.6 & 23.9 & 16.4 & 10.2 & 8.0 & 5.7 & 5.0 & 4.8 \\
    Vietnamese & 56.1 & 30.8 & 24.9 & 18.9 & 15.4 & 12.9 & 8.9 & 9.0 \\
    Cantonese & - & - & - & - & 13.8 & 11.1 & 10.0 & 9.4 \\
    Hindi & 105.4 & 74.1 & 61.9 & 45.4 & 29.9 & 26.2 & 25.4 & 24.9 \\
    Urdu & 56.5 & 41.6 & 33.3 & 25.8 & 15.0 & 11.7 & 10.6 & 10.3 \\
    Malay & 61.8 & 41.0 & 32.9 & 26.5 & 21.2 & 17.5 & 15.5 & 15.3 \\
    Uzbek & 131.6 & 125.4 & 115.9 & 86.3 & 21.9 & 16.3 & 14.7 & 13.9 \\
    Arabic & 59.8 & 40.9 & 30.6 & 21.7 & 28.1 & 18.7 & 16.6 & 15.1 \\
    Persian & 90.9 & 62.9 & 42.7 & 28.1 & 19.3 & 14.1 & 12.9 & 12.2 \\
    Bengali & 115.7 & 136.1 & 127.0 & 71.0 & 23.4 & 17.9 & 16.1 & 15.4 \\
    Tamil & 100.3 & 94.7 & 92.4 & 85.6 & 59.3 & 46.3 & 44.0 & 43.0 \\
    Telugu & 120.6 & 108.7 & 119.6 & 94.6 & 58.5 & 46.2 & 44.0 & 42.5 \\
    Uighur & - & - & - & - & 43.0 & 31.8 & 29.0 & 28.4 \\
    Gujarati & 107.6 & 109.5 & 108.4 & 74.2 & 39.2 & 31.4 & 29.1 & 28.0 \\
    Myanmar & 100.4 & 101.5 & 100.4 & 98.7 & 15.7 & 10.4 & 9.1 & 8.4 \\
    Tagalog & 64.4 & 38.4 & 26.2 & 19.6 & 20.4 & 15.5 & 13.5 & 12.7 \\
    Kazakh & 107.9 & 87.4 & 71.1 & 51.6 & 33.9 & 23.6 & 20.0 & 18.6 \\
    Oriya & - & - & - & - & 34.6 & 26.8 & 24.1 & 22.6\\
    Nepali & 134.6 & 116.0 & 101.7 & 88.8 & 26.7 & 20.7 & 18.8 & 18.0 \\
    Mongolian & 114.8 & 140.8 & 106.5 & 84.3 & 29.4 & 18.4 & 15.6 & 13.5 \\
    Khmer & 110.9 & 109.8 & 114.7 & 100.0 & 42.5 & 34.1 & 31.7 & 30.4 \\
    Javanese & 87.7 & 100.0 & 71.9 & 60.2 & 7.5 & 5.5 & 4.9 & 4.7 \\
    Lao & 103.7 & 101.5 & 102.3 & 102.7 & 14.5 & 11.5 & 10.6 & 10.3 \\
    Sinhala & 117.3 & 127.5 & 128.3 & 108.9 & 40.8 & 30.3 & 26.0 & 24.7 \\
    Filipino & - & - & - & - & 16.5 & 10.5 & 8.6 & 7.6 \\
    Pashto & 99.7 & 94.3 & 105.6 & 88.0 & 41.5 & 33.6 & 30.4 & 29.6 \\
    Punjabi & 105.4 & 122.3 & 112.4 & 83.6 & 49.3 & 41.2 & 38.3 & 37.4 \\
    Kashmiri & - & - & - & - & 61.4 & 53.8 & 51.2 & 49.4 \\
    Tajik & 120.9 & 93.4 & 83.4 & 81.0 & 36.9 & 23.9 & 20.4 & 19.6 \\
    Sundanese & 81.1 & 68.9 & 64.5 & 59.5 & 21.6 & 15.5 & 13.7 & 12.3 \\
    Marathi & 119.9 & 106.8 & 99.1 & 78.8 & 47.9 & 32.1 & 28.2 & 24.9 \\
    Kyrgyz & - & - & - & - & 89.8 & 75.4 & 72.0 & 75.7 \\
    Azerbaijani & 97.8 & 80.2 & 67.2 & 55.1 & 75.7 & 59.1 & 52.5 & 47.7 \\
    \bottomrule
  \end{tabular}
  }
\end{table*}

\begin{table*}[h]
  \caption{WER (\%) on CommonVoice17 dataset.}
  \label{tab:language_comparison}
  \centering
  \scalebox{1}{
  \begin{tabular}{l|ccccc|cccc}
    \toprule
     & \multicolumn{5}{c}{\textbf{Whisper}} & \multicolumn{4}{c}{\textbf{Dolphin}} \\
    \textbf{Models} & \textbf{base} & \textbf{small} & \textbf{medium} & \textbf{large-v1} & \textbf{large-v3} & \textbf{base} & \textbf{small} & \textbf{medium} & \textbf{large} \\
    \midrule
    Chinese    & 48.1  & 32.7  & 26.3  & 29.4  & 15.4  & 13.1  & 9.6   & 11.2 & 8.2  \\
    Japanese   & 37.8  & 23.8  & 17.9  & 17.9  & 15.2  & 18.2  & 16.8  & 14.1 & 13.8 \\
    Thai       & 34.3  & 19.5  & 12.8  & 10.3  & 7.1   & 6.2   & 4.8   & 4.4  & 4.2  \\
    Russian    & 34.4  & 18.2  & 11.7  & 10.3  & 7.4   & 22.8  & 13.6  & 13.0 & 9.8  \\
    Korean     & 24.4  & 14.3  & 10.2  & 8.9   & 11.9  & 12.1  & 8.0   & 7.1  & 6.5  \\
    Indonesian & 45.0  & 22.2  & 13.6  & 12.3  & 7.8   & 14.0  & 9.7   & 10.2 & 8.7  \\
    Vietnamese & 51.8  & 31.1  & 24.3  & 20.7  & 14.8  & 20.2  & 16.3  & 13.4 & 12.1 \\
    Cantonese  & -     & -     & -     & -     & -     & 16.9  & 9.0   & 7.9  & 6.7  \\
    Hindi      & 107.2 & 64.2  & 49.2  & 47.5  & 35.2  & 21.2  & 15.9  & 14.2 & 13.4 \\
    Urdu       & 63.2  & 43.6  & 33.3  & 33.6  & 25.1  & 27.7  & 23.7  & 21.1 & 21.1 \\
    Uzbek      & 119.5 & 123.1 & 119.5 & 95.8  & 90.8  & 28.9  & 19.1  & 18.1 & 16.2 \\
    Arabic     & 82.5  & 55.0  & 43.6  & 41.7  & 33.5  & 51.3  & 40.1  & 37.0 & 35.8 \\
    Persian    & 103.4 & 77.1  & 62.2  & 51.2  & 37.4  & 32.0  & 24.2  & 22.4 & 21.2 \\
    Bengali    & 117.8 & 130.0 & 126.0 & 121.5 & 77.0  & 31.9  & 22.4  & 19.4 & 17.6 \\
    Tamil      & 90.3  & 69.1  & 58.7  & 54.6  & 53.1  & 49.0  & 39.5  & 36.1 & 34.5 \\
    Telugu     & 167.6 & 163.7 & 187.4 & 124.7 & 80.2  & 69.8  & 62.1  & 62.1 & 55.0 \\
    Uighur     & -     & -     & -     & -     & -     & 31.3  & 20.0  & 17.4 & 15.4 \\
    Kazakh     & 112.4 & 85.9  & 69.1  & 64.5  & 51.7  & 55.4  & 37.9  & 34.4 & 30.4 \\
    Oriya      & -     & -     & -     & -     & -     & 46.3  & 38.6  & 35.5 & 33.9 \\
    Nepali     & 112.5 & 109.7 & 98.5  & 106.6 & 84.6  & 42.3  & 32.3  & 31.3 & 30.5 \\
    Mongolian  & 111.1 & 134.2 & 109.2 & 105.7 & 88.4  & 44.2  & 28.8  & 24.3 & 21.7 \\
    Lao        & 102.1 & 102.0 & 102.2 & 101.8 & 102.5 & 19.9  & 10.8  & 14.9 & 13.7 \\
    Pashto     & 99.2  & 94.4  & 114.5 & 104.5 & 89.3  & 57.9  & 46.2  & 43.2 & 39.4 \\
    Punjabi    & 105.2 & 137.4 & 129.6 & 101.2 & 69.5  & 41.6  & 32.7  & 29.9 & 25.4 \\
    Kabyle     & -     & -     & -     & -     & -     & 65.2  & 45.8  & 38.6 & 35.7 \\
    Bashkir    & 122.0 & 120.5 & 113.0 & 105.6 & 103.5 & 36.0  & 23.4  & 17.7 & 15.6 \\
    Marathi    & 125.6 & 116.6 & 115.8 & 95.1  & 80.3  & 55.4  & 38.8  & 33.6 & 31.2 \\
    Kyrgyz     & -     & -     & -     & -     & -     & 64.5  & 42.5  & 35.1 & 33.7 \\
    Azerbaijani& 79.4  & 61.5  & 42.1  & 35.3  & 23.0  & 82.1  & 61.9  & 56.4 & 54.0 \\
    \bottomrule
  \end{tabular}
  }
\end{table*}

\begin{table*}[h]
  \caption{WER (\%) on Fleurs dataset.}
  \label{tab:fleurs results}
  \centering
  \scalebox{1}{
  \begin{tabular}{l|ccccc|cccc}
    \toprule
     & \multicolumn{5}{c}{\textbf{Whisper}} & \multicolumn{4}{c}{\textbf{Dolphin}} \\
    \textbf{Models} & \textbf{base} & \textbf{small} & \textbf{medium} & \textbf{large-v1} & \textbf{large-v3} & \textbf{base} & \textbf{small} & \textbf{medium} & \textbf{large} \\
    \midrule
Chinese & 31.3 & 18.1 & 10.7 & 16.4 & 7.2 & 6.5 & 4.6 & 4.1 & 4.0 \\
Japanese & 25.0 & 12.7 & 7.6 & 7.1 & 4.8 & 7.4 & 5.5 & 4.7 & 4.6 \\
Thai & 40.4 & 25.6 & 18.8 & 15.9 & 19.1 & 12.0 & 11.3 & 11.0 & 10.8 \\
Russian & 23.5 & 12.9 & 9.7 & 7.2 & 5.4 & 20.0 & 13.9 & 13.6 & 12.2 \\
Korean & 10.3 & 5.3 & 3.5 & 3.1 & 2.6 & 4.8 & 3.3 & 2.7 & 2.5 \\
Indonesian & 39.7 & 18.1 & 11.5 & 9 & 6.5 & 15.2 & 12.8 & 12.3 & 11.7 \\
Vietnamese & 42.4 & 22.3 & 13.9 & 11.4 & 8.9 & 17.3 & 13.8 & 12.8 & 13.0 \\
Cantonese & - & - & - & - & - & 13.6 & 9.8 & 8.9 & 8.4 \\
Hindi & 107.6 & 58.9 & 43.8 & 44.9 & 16.5 & 18.3 & 14.4 & 13.1 & 12.5 \\
Urdu & 54.8 & 40 & 29.5 & 27.1 & 21.5 & 24.9 & 19.3 & 18.6 & 17.0 \\
Malay & 41.9 & 21.9 & 13.7 & 11.6 & 8.2 & 17.6 & 12.9 & 11.7 & 11.4 \\
Uzbek & 115.6 & 109.9 & 109.9 & 96 & 87.1 & 36.3 & 27.6 & 26.5 & 25.1 \\
Arabic & 50.8 & 29.6 & 18.3 & 15.9 & 10.5 & 22.4 & 14.2 & 12.1 & 11.3 \\
Persian & 88.2 & 58.2 & 45.9 & 39.2 & 30.7 & 29.1 & 24.3 & 22.8 & 23.0 \\
Bengali & 115.4 & 115.5 & 111 & 104.9 & 47.9 & 25.7 & 20.0 & 18.3 & 17.1 \\
Tamil & 99.1 & 83.2 & 65.9 & 62.7 & 29.7 & 47.0 & 38.0 & 35.5 & 33.1 \\
Telugu & 118.9 & 111.3 & 107 & 101.3 & 39.2 & 47.4 & 37.8 & 34.8 & 32.1 \\
Gujarati & 109.7 & 111.5 & 110.9 & 109 & 41.6 & 42.6 & 35.4 & 32.2 & 32.4 \\
Myanmar & 100.1 & 100.8 & 100 & 100.6 & 100.2 & 18.1 & 13.4 & 11.7 & 11.0 \\
Kazakh & 99.4 & 75.1 & 54 & 48.9 & 33.3 & 31.5 & 21.1 & 18.3 & 16.7 \\
Oriya & - & - & - & - & - & 40.8 & 33.8 & 27.4 & 26.8\\
Nepali & 137.2 & 120.2 & 108.3 & 110 & 41.0 & 41.1 & 36.6 & 29.2 & 28.4 \\
Mongolian & 106.3 & 114.7 & 104.3 & 102.6 & 86.0 & 43.2 & 29.6 & 25.4 & 22.8 \\
Khmer & 101.7 & 101 & 100.9 & 99.4 & 85.6 & 20.9 & 14.8 & 12.9 & 12.1 \\
Javanese & 93.3 & 100.2 & 72 & 97.6 & 65.7 & 27.7 & 22.2 & 20.2 & 20.1 \\
Lao & 104.7 & 103 & 103.4 & 103.2 & 106.0 & 26.1 & 21.5 & 21.3 & 19.3 \\
Filipino & - & - & - & - & - & 21.6 & 16.1 & 14.8 & 13.9 \\
Pashto & 105.2 & 93.6 & 103 & 98.6 & 88.6 & 55.2 & 47.3 & 45.7 & 44.8 \\
Punjabi & 105.1 & 108.3 & 106.4 & 101.6 & 46.7 & 45.5 & 34.6 & 33.6 & 29.2 \\
Tajik & 122.8 & 88.6 & 76.9 & 78.1 & 81.2 & 34.5 & 20.8 & 20.2 & 17.2 \\
Marathi & 115.3 & 110.8 & 105.7 & 97.1 & 35.3 & 54.7 & 38.5 & 33.5 & 31.1 \\
Kyrgyz & - & - & - & - & - & 72.5 & 52.7 & 69.9 & 58.3 \\
Azerbaijani & 80.9 & 52.2 & 35.7 & 70.2 & 21.6 & 88.4 & 58.1 & 53.8 & 47.0 \\
    \bottomrule
  \end{tabular}
  }
\end{table*}

\onecolumn
\section{Language Region Code}
\label{sec:appendixb}
\captionof{table}{Language Region Code.}
\tablefirsthead{
\toprule
\textbf{Language-Region Code} & \textbf{Name} \\
\midrule
}
\tablehead{
\multicolumn{2}{c}{Table \thetable\ \textit{(Continued)}} \\
\toprule
\textbf{Language-Region Code} & \textbf{Name} \\
\midrule
}
\tabletail{
\bottomrule
\multicolumn{2}{r}{{Continued on next page}} \\
}
\tablelasttail{
\bottomrule
}
\centering
\begin{supertabular}{l|l}
zh-CN & Chinese (Mandarin) \\
zh-TW & Chinese (Taiwan) \\
zh-WU & Chinese (Wuyu) \\
zh-SICHUAN & Chinese (Sichuan) \\
zh-SHANXI & Chinese (Shanxi) \\
zh-ANHUI & Chinese (Anhui) \\
zh-TIANJIN & Chinese (Tianjin) \\
zh-NINGXIA & Chinese (Ningxia) \\
zh-SHAANXI & Chinese (Shaanxi) \\
zh-HEBEI & Chinese (Hebei) \\
zh-SHANDONG & Chinese (Shandong) \\
zh-GUANGDONG & Chinese (Guangdong) \\
zh-SHANGHAI & Chinese (Shanghai) \\
zh-HUBEI & Chinese (Hubei) \\
zh-LIAONING & Chinese (Liaoning) \\
zh-GANSU & Chinese (Gansu) \\
zh-FUJIAN & Chinese (Fujian) \\
zh-HUNAN & Chinese (Hunan) \\
zh-HENAN & Chinese (Henan) \\
zh-YUNNAN & Chinese (Yunnan) \\
zh-MINNAN & Chinese (Minnan) \\
zh-WENZHOU & Chinese (Wenzhou) \\
ja-JP & Japanese \\
th-TH & Thai \\
ru-RU & Russian \\
ko-KR & Korean \\
id-ID & Indonesian \\
vi-VN & Vietnamese \\
ct-NULL & Yue (Unknown) \\
ct-HK & Yue (Hongkong) \\
ct-GZ & Yue (Guangdong) \\
hi-IN & Hindi \\
ur-IN & Urdu \\
ur-PK & Urdu (Islamic Republic of Pakistan) \\
ms-MY & Malay \\
uz-UZ & Uzbek \\
ar-MA & Arabic (Morocco) \\
ar-GLA & Arabic \\
ar-SA & Arabic (Saudi Arabia) \\
ar-EG & Arabic (Egypt) \\
ar-KW & Arabic (Kuwait) \\
ar-LY & Arabic (Libya) \\
ar-JO & Arabic (Jordan) \\
ar-AE & Arabic (U.A.E.) \\
ar-LVT & Arabic (Levant) \\
fa-IR & Persian \\
bn-BD & Bengali \\
ta-SG & Tamil (Singaporean) \\
ta-LK & Tamil (Sri Lankan) \\
ta-IN & Tamil (India) \\
ta-MY & Tamil (Malaysia) \\
te-IN & Telugu \\
ug-NULL & Uighur \\
ug-CN & Uighur \\
gu-IN & Gujarati \\
my-MM & Burmese \\
tl-PH & Tagalog \\
kk-KZ & Kazakh \\
or-IN & Oriya / Odia \\
ne-NP & Nepali \\
mn-MN & Mongolian \\
km-KH & Khmer \\
jv-ID & Javanese \\
lo-LA & Lao \\
si-LK & Sinhala \\
fil-PH & Filipino \\
ps-AF & Pushto \\
pa-IN & Panjabi \\
kab-NULL & Kabyle \\
ba-NULL & Bashkir \\
ks-IN & Kashmiri \\
tg-TJ & Tajik \\
su-ID & Sundanese \\
mr-IN & Marathi \\
ky-KG & Kirghiz \\
az-AZ & Azerbaijani \\ 
\end{supertabular}

\end{document}